\documentclass{article} 
\usepackage{iclr2020_conference,times}

\usepackage{amsmath,amsfonts,bm}

\def\eqref#1{equation~\ref{#1}}

\def\1{\bm{1}}

\def\eps{{\epsilon}}

\DeclareMathAlphabet{\mathsfit}{\encodingdefault}{\sfdefault}{m}{sl}
\SetMathAlphabet{\mathsfit}{bold}{\encodingdefault}{\sfdefault}{bx}{n}

\usepackage{hyperref}
\usepackage{url}
\usepackage{graphicx}
\usepackage{subcaption}
\usepackage{algorithm}
\usepackage{algpseudocode}

\title{Instance adaptive adversarial training: \\ Improved accuracy tradeoffs in neural nets}

\author{
Yogesh Balaji$^{1, 2}$\thanks{Work done during an internship at Facebook AI Research} \hfill
Tom Goldstein $^{1, 2}$ \hfill
Judy Hoffman $^{1, 3}$
\\$^{1}$\normalfont{Facebook AI Research} \quad
$^{2}$\normalfont{University of Maryland} \quad 
$^{3}$\normalfont{Georgia Institute of Technology} \\
} 

\newcommand{\bx}{\mathbf{x}}
\newcommand{\by}{\mathbf{y}}

\newcommand{\cD}{\mathcal{D}}
\newcommand{\RR}{\mathbb{R}}

\usepackage{xcolor}

\iclrfinalcopy 
\begin{document}

\maketitle

\begin{abstract}
Adversarial training is by far the most successful strategy for improving robustness of neural networks to adversarial attacks. Despite its success as a defense mechanism, adversarial training fails to generalize well to unperturbed test set. We hypothesize that this poor generalization is a consequence of adversarial training with uniform perturbation radius around every training sample. Samples close to decision boundary can be morphed into a different class under a small perturbation budget, and enforcing large margins around these samples produce poor decision boundaries that generalize poorly. Motivated by this hypothesis, we propose instance adaptive adversarial training -- a technique that enforces sample-specific perturbation margins around every training sample. We show that using our approach, test accuracy on unperturbed samples improve with a marginal drop in robustness. Extensive experiments on CIFAR-10, CIFAR-100 and Imagenet datasets demonstrate the effectiveness of our proposed approach.
\end{abstract}


\section{Introduction}\label{sec:intro}

A key challenge when deploying neural networks in safety-critical applications is their poor stability to input perturbations. Extremely tiny perturbations to network inputs may be imperceptible to the human eye, and yet cause major changes to outputs. One of the most effective and widely used methods for hardening networks to small perturbations is ``adversarial training"~\citep{madry2018towards}, in which a network is trained using adversarially perturbed samples with a fixed perturbation size.  By doing so, adversarial training typically tries to enforce that the output of a neural network remains nearly constant within an $\ell_p$ ball of every training input.

Despite its ability to increase robustness, adversarial training suffers from poor accuracy on clean (natural) test inputs.  The drop in clean accuracy can be as high as $~10\%$ on CIFAR-10, and $~15\%$ on Imagenet~\citep{madry2018towards, Xie_2019_feature}, making robust models undesirable in some industrial settings. The consistently poor performance of robust models on clean data has lead to the line of thought that there may be a fundamental trade-off between robustness and accuracy~\citep{zhang2019TRADES, Tsipras2019robustness}, and recent theoretical results characterized this tradeoff \citep{fawzi2018adversarial,shafahi2018adversarial,mahloujifar2019curse}.

\begin{figure*}
\centering
\includegraphics[width=\textwidth]{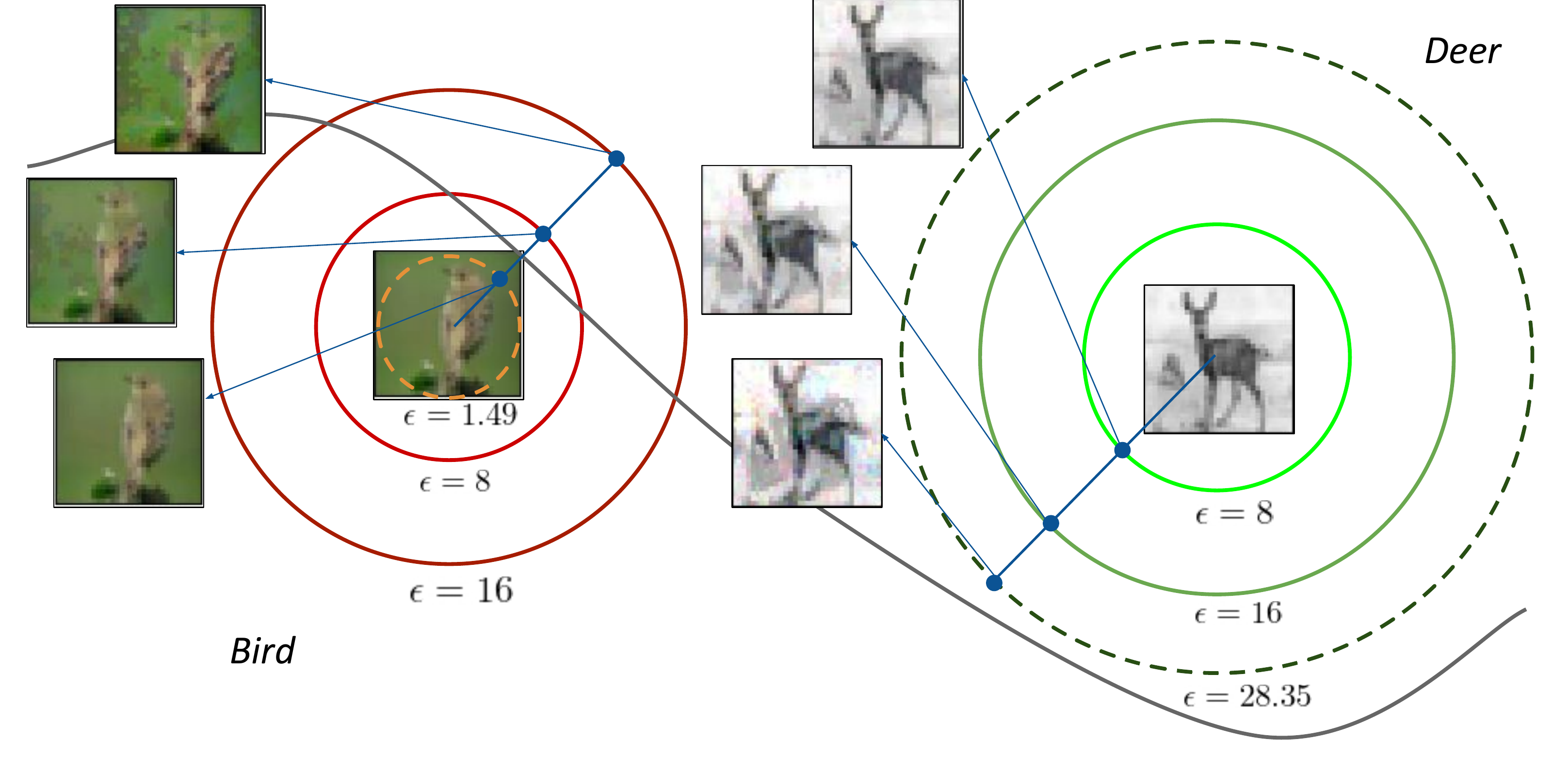}
\caption{Overview of instance adaptive adversarial training. Samples close to the decision boundary (bird on the left) have nearby samples from a different class (deer) within a small $L_p$ ball, making the constraints imposed by PGD-8 / PGD-16 adversarial training infeasible. Samples far from the decision boundary (deer on the right) can withstand large perturbations well beyond $\epsilon=8$. Our adaptive adversarial training correctly assigns the perturbation radius (shown in dotted line) so that samples within each $L_p$ ball maintain the same class.}
\label{fig:title_fig} 
\end{figure*}

In this work, we aim to understand and optimize the tradeoff between robustness and clean accuracy. More concretely, our objective is to improve the clean accuracy of adversarial training for a chosen level of adversarial robustness.   Our method is inspired by the observation that the constraints enforced by adversarial training are {\em infeasible}; for commonly used values of $\epsilon,$ it is not possible to achieve label consistency within an $\epsilon$-ball of each input image because the balls around images of different classes overlap.  This is illustrated on the left of Figure~\ref{fig:title_fig}, which shows that the $\epsilon$-ball around a  ``bird'' (from the CIFAR-10 training set) contains images of class ``deer'' (that do not appear in the training set).  If adversarial training were successful at enforcing label stability in an $\epsilon=8$ ball around the ``bird'' training image, doing so would come at the {\em unavoidable} cost of misclassifying the nearby ``deer'' images that come along at test time.   At the same time, when training images lie far from the decision boundary (eg., the deer image on the right in Fig~\ref{fig:title_fig}), it is possible to enforce stability with large $\epsilon$ with no compromise in clean accuracy.  When adversarial training on CIFAR-10, we see that $\epsilon=8$ is too large for some images, causing accuracy loss, while being unnecessarily small for others, leading to sub-optimal robustness.

The above observation naturally motivates adversarial training with {\em instance adaptive} perturbation radii that are customized to each training image. By choosing larger robustness radii at locations where class manifolds are far apart, and smaller radii at locations where class manifolds are close together, we get high adversarial robustness where possible while minimizing the clean accuracy loss that comes from enforcing overly-stringent constraints on images that lie near class boundaries. As a result, instance adaptive training significantly improves the tradeoff between accuracy and robustness, breaking through the pareto frontier achieved by standard adversarial training. 
Additionally, we show that the learned instance-specific perturbation radii are interpretable; samples with small radii are often ambiguous and have nearby images of another class, while images with large radii have unambiguous class labels that are difficult to manipulate. 

Parallel to our work, we found that \cite{ding2018max} uses adaptive margins in a max-margin framework for adversarial training. Their work focuses on improving the adversarial robustness, which differs from our goal of understanding and improving the robustness-accuracy tradeoff. Moreover, our algorithm for choosing adaptive margins significantly differs from that of \cite{ding2018max}.

\section{Background}\label{sec:background}

Adversarial attacks are data items containing small perturbations that cause misclassification in neural network classifiers~\citep{szegedy2014intriguing}. Popular methods for crafting attacks include the fast gradient sign method (FGSM)~\citep{goodfellow2015FGSM} which is a one-step gradient attack, projected gradient descent (PGD)~\citep{madry2018towards} which is a multi-step extension of FGSM, the C/W attack~\citep{carlini2017towards}, DeepFool~\citep{Moosavi2016deepfool}, and many more. All these methods use the gradient of the loss function with respect to inputs to construct additive perturbations with a norm-constraint. Alternative attack metrics include spatial transformer attacks~\citep{xiao2018spatially}, attacks based on Wasserstein distance in pixel space~\citep{pmlr-v97-wong19a}, etc.

Defending against adversarial attacks is a crucial problem in machine learning. Many early defenses~\citep{buckman2018thermometer, samangouei2018defensegan, dhillon2018stochastic}, were broken by strong attacks. Fortunately, \textit{adversarially training} is one defense strategy that remains fairly resistant to most existing attacks.

Let $\cD = \{ (\bx_i, y_i) \}_{i=1}^{n}$ denote the set of training samples in the input dataset. In this paper, we focus on classification problems, hence, $y_i \in \{1, 2, \hdots N_c \}$, where $N_c$ denotes the number of classes. Let $f_{\theta}(\bx) : \RR^{c \times m \times n} \to \RR^{N_c}$ denote a neural network model parameterized by $\theta$. Classifiers are often trained by minimizing the cross entropy loss given by
\begin{align*}
    \min_{\theta} \frac{1}{N} \sum_{(\bx_i, y_i) \sim \cD} -\tilde{\by_i} \big[ \log(f_{\theta}(\bx_i)) \big]
\end{align*}
where $\tilde{\by_i}$ is the one-hot vector corresponding to the label $y_i$.
In adversarial training, instead of optimizing the neural network over the clean training set, we use the adversarially perturbed training set. Mathematically, this can be written as the following \textit{min-max} problem
\begin{align}\label{eq:PGD}
    \min_{\theta} \max_{ \|\delta_i\|_{\infty} \le \epsilon} \frac{1}{N} \sum_{(\bx_i, y_i) \sim \cD} -\tilde{\by_i} \big[ \log(f_{\theta}(\bx_i) + \delta_i) \big]
\end{align}
This problem is solved by an alternating stochastic method that takes minimization steps for $\theta,$ followed by maximization steps that approximately solve the inner problem using  $k$ steps of PGD. For more details, refer to \cite{madry2018towards}.


\begin{algorithm}
\caption{Adaptive adversarial training algorithm}
\label{alg:adaptive_adv}
\begin{algorithmic}[1]
\Require $N_{iter}$: Number of training iterations, $N_{warm}$: Warmup period
\Require $PGD_{k}(\bx, y, \epsilon): $ Function to generate PGD-$k$ adversarial samples with $\epsilon$ norm-bound
\Require $\epsilon_{w}$: $\epsilon$ used in warmup
\For{$t$ in $1:N_{iter}$}
    \State Sample a batch of training samples $\{ (\bx_i, y_i) \}_{i=1}^{N_{batch}} \sim \cD$
    \If {$t < N_{warm}$}
        \State $\epsilon_i = \epsilon_{w}$
    \Else
        \State Choose $\epsilon_i$ using Alg~\ref{alg:epsilon_select}
    \EndIf
    \State $\bx^{adv}_{i} = PGD(\bx_i, y_i, \epsilon_{i})$
    \State $S_{+} = \{ i| f(\bx_i) \text{ is correctly classified as } y_i \}$
    \State $S_{-} = \{ i| f(\bx_i) \text{ is incorrectly classified as } y_i \}$
    \State $\min_{\theta} \frac{1}{N_{batch}} \Big[ \sum_{i \in S_{+}} L_{cls}(\bx_{i}^{adv}, y^i) + \sum_{i \in S_{-}} L_{cls}(\bx_i, y_i) \Big]$
\EndFor
\end{algorithmic}
\end{algorithm}

\section{Instance Adaptive Adversarial Training}\label{sec:method}

\begin{figure*}[t!]
    \centering
    \begin{subfigure}[t]{0.5\textwidth}
        \centering
        \includegraphics[height=2.5in]{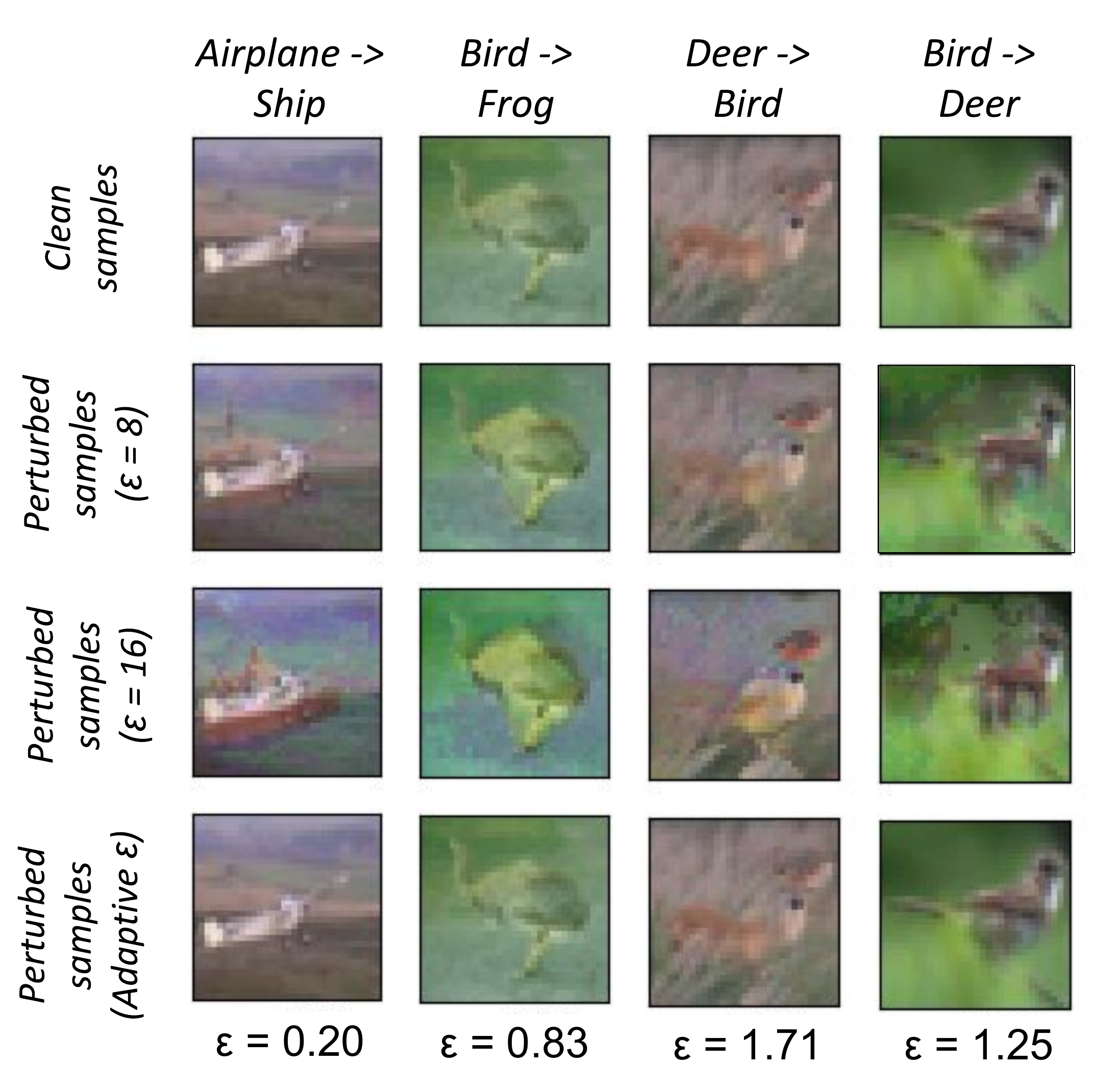}
        \caption{Samples from bottom $1 \%$ $\epsilon$}
    \end{subfigure}%
    ~ 
    \begin{subfigure}[t]{0.47\textwidth}
        \centering
        \includegraphics[height=2.5in]{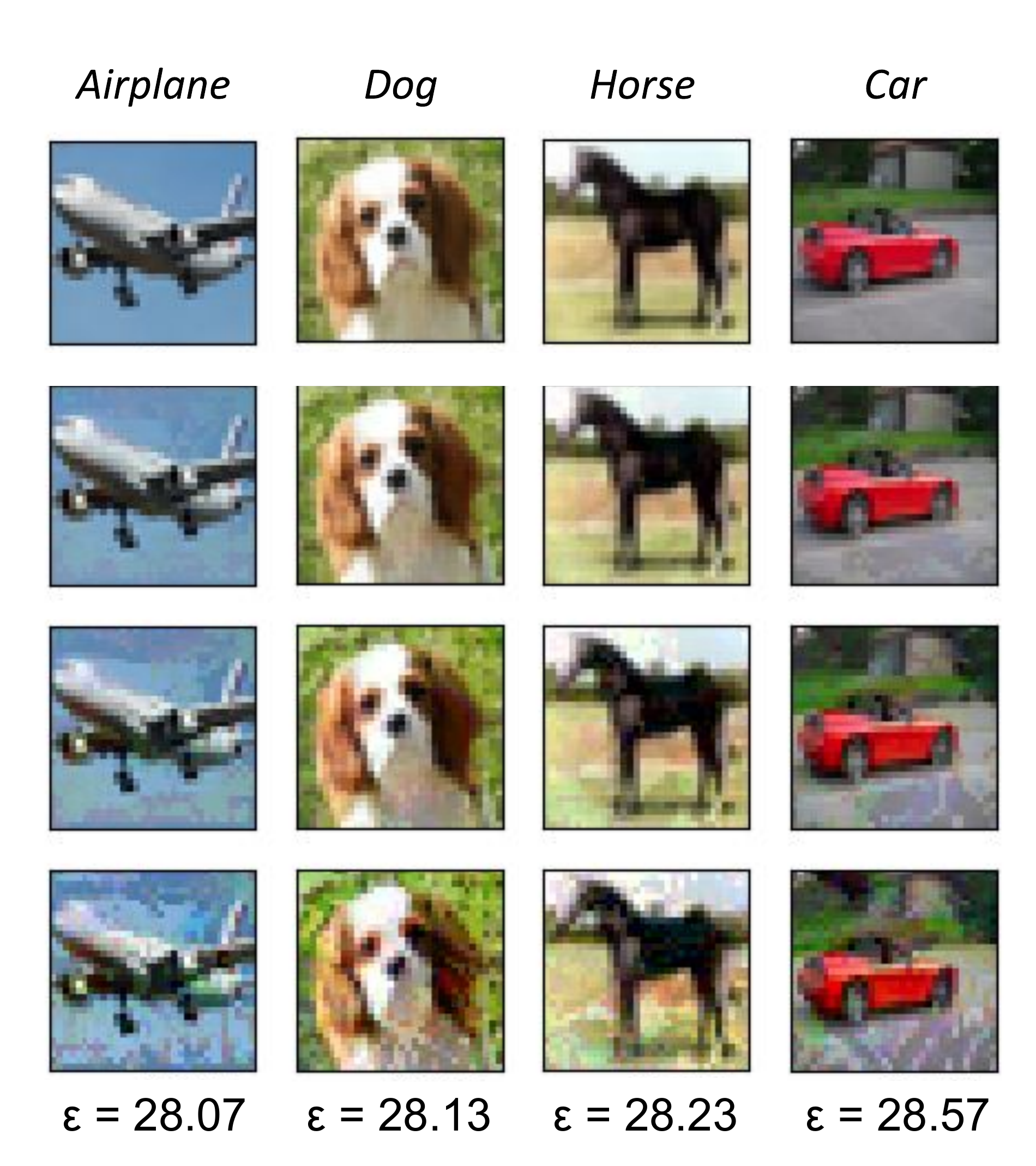}
        \caption{Samples from top $1 \%$ $\epsilon$}
    \end{subfigure}
    \caption{Visualizing training samples and their perturbations. The left panel shows samples that are assigned small $\epsilon$ (displayed below images) during adaptive training.  These images are close to class boundaries, and change class when perturbed with $\epsilon \ge 8$. The right panel show images that are assigned large $\epsilon.$  These lie far from the decision boundary, and retain class information even with very large perturbations. All $\epsilon$ live in the range $[0, 255]$}
    \label{fig:sample_vis_main}
\end{figure*}

To remedy the shortcomings of uniform perturbation radius in adversarial training (Section~\ref{sec:intro}), we propose {\em Instance Adaptive Adversarial Training} (IAAT), which solves the following optimization: 
\begin{align}\label{eq:PGD-adaptive}
    \min_{\theta} \max_{ \|\delta_i\|_{\infty} < \epsilon_{i}} \frac{1}{N} \sum_{(\bx_i, y_i) \sim \cD} -\tilde{\by_i} \big[ \log(f_{\theta}(\bx_i) + \delta_i) \big]
\end{align}
Like vanilla adversarial training, we solve this by sampling mini-batches of images $\{\bx_i\}$, crafting adversarial perturbations $\{\delta_i\}$ of size at most $\{\epsilon_i\}$, and then updating the network model using the perturbed images.

The proposed algorithm is distinctive in that it uses a different $\epsilon_i$ for each image $\bx_i.$
Ideally, we would choose each $\epsilon_{i}$ to be as large as possible without finding images of a different class within the $\epsilon_{i}$-ball around $\bx_i.$  Since we have no a-priori knowledge of what this radius is, we use a simple heuristic to update $\epsilon_{i}$ after each epoch. After crafting a perturbation for $\bx_i,$ we check if the perturbed image was a successful adversarial example.  If PGD succeeded in finding an image with a different class label, then $\epsilon_{i}$ is too big, so we replace $\epsilon_{i} \gets \epsilon_{i} - \gamma$.  If PGD failed, then we set $\epsilon_{i} \gets \epsilon_{i} + \gamma$.

Since the network is randomly initialized at the start of training, random predictions are made, and this causes $\{\epsilon_i\}$ to shrink rapidly. For this reason, we begin with a warmup period of a few (usually 10 epochs for CIFAR-10/100) epochs where adversarial training is performed using uniform $\epsilon$ for every sample. After the warmup period ends, we perform instance adaptive adversarial training.

A detailed training algorithm is provided in Alg.~\ref{alg:adaptive_adv}.

\begin{algorithm}
\caption{$\epsilon$ selection algorithm}
\label{alg:epsilon_select}
\begin{algorithmic}[1]
\Require $i$: Sample index, $j$: Epoch index
\Require $\beta$: Smoothing constant, $\gamma$: Discretization for $\epsilon$ search.
\State Set $\epsilon_1 = \bm{\epsilon}_{mem}[j-1, i] + \gamma$
\State Set $\epsilon_2 = \bm{\epsilon}_{mem}[j-1, i]$
\State Set $\epsilon_3 = \bm{\epsilon}_{mem}[j-1, i] - \gamma$
\If{$f_{\theta}(PGD_{k}(\bx_i, y_i, \epsilon_1))$ predicts as $y_i$}
    \State Set $\epsilon_i = \epsilon_1$
\ElsIf{$f_{\theta}(PGD_{k}(\bx_i, y_i, \epsilon_2))$ predicts as $y_i$}
    \State Set $\epsilon_i = \epsilon_2$
\Else
    \State Set $\epsilon_i = \epsilon_3$
\EndIf
\State $\epsilon_{i} \gets (1-\beta) \bm{\epsilon}_{mem}[j-1, i] + \beta \epsilon_i $
\State Update $\bm{\epsilon}_{mem}[j, i] \gets \epsilon_i$
\State Return $\epsilon_i$
\end{algorithmic}
\end{algorithm}


\section{Experiments}\label{sec:experiments}
 To evaluate the robustness and generalization of our models, we report the following metrics: (1) test accuracy of unperturbed (natural) test samples, (2) adversarial accuracy of white-box PGD attacks, (3) adversarial accuracy of transfer attacks and (4) accuracy of test samples under common image corruptions \citep{hendrycks2018benchmarking}. Following the protocol introduced in \cite{hendrycks2018benchmarking}, we do not train our models on any image corruptions. 

\subsection{CIFAR}
On CIFAR-10 and CIFAR-100 datasets, we perform experiments on Resnet-18 and WideRenset-32-10 models following ~\citep{madry2018towards, zhang2019TRADES}. All models are trained on PGD-$10$ attacks i.e., $10$ steps of PGD iterations are used for crafting adversarial attacks during training. In the whitebox setting, models are evaluated on: (1) PGD-$10$ attacks with $5$ random restarts, (2) PGD-$100$ attacks with $5$ random restarts, and (3) PGD-$1000$ attacks with $2$ random restarts. For transfer attacks, an independent copy of the model is trained using the same training algorithm and hyper-parameter settings, and PGD-$1000$ adversarial attacks with $2$ random restarts are crafted on the surrogate model. For image corruptions, following \citep{hendrycks2018benchmarking}, we report average classification accuracy on $19$ image corruptions. 

\begin{figure}
  \centering
  \begin{subfigure}[b]{0.45\textwidth}
    \includegraphics[width=\textwidth]{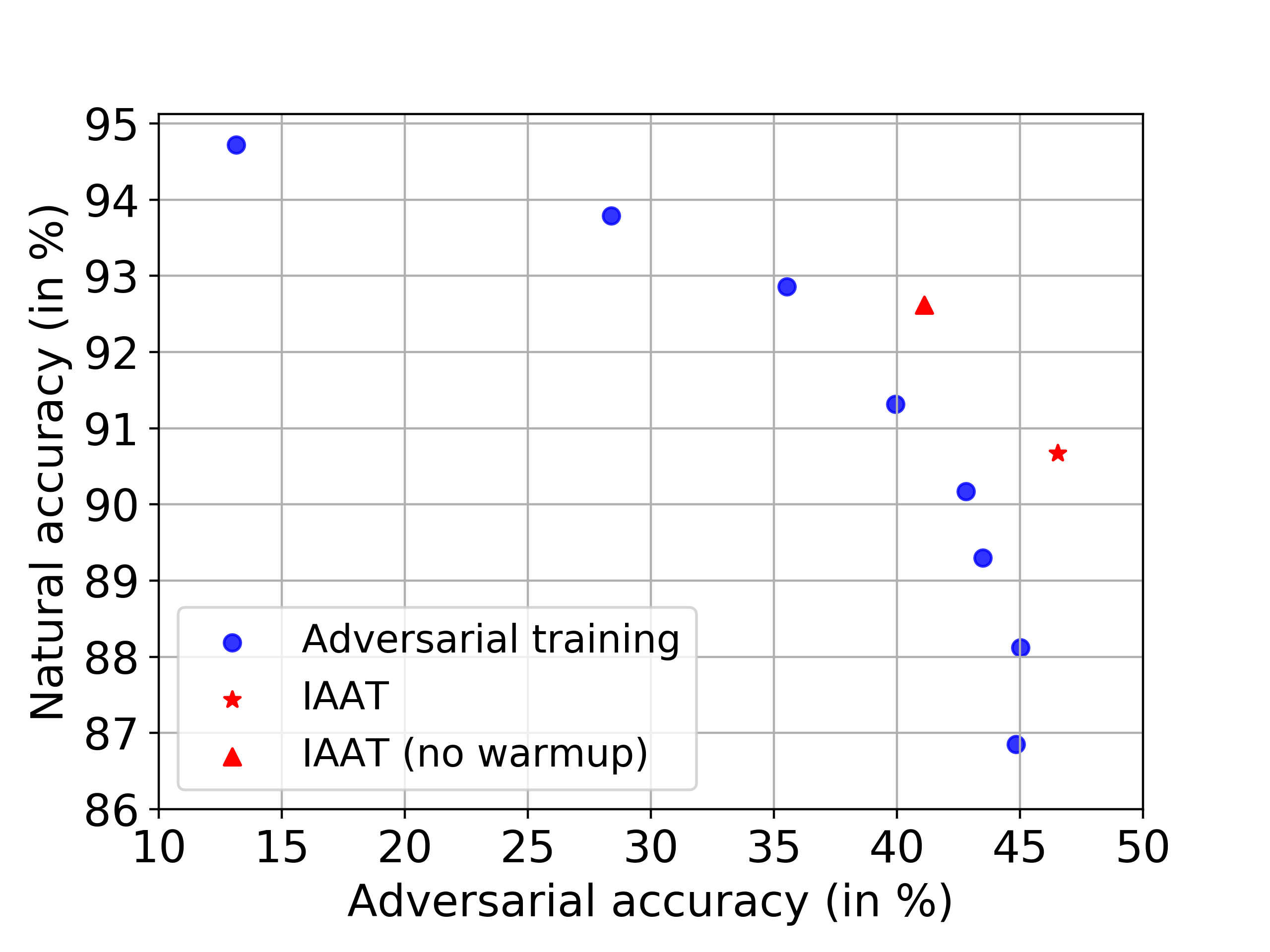}
    \caption{CIFAR-10}
    \label{fig:tradeoff_CIFAR10}
  \end{subfigure}
  \begin{subfigure}[b]{0.45\textwidth}
    \includegraphics[width=\textwidth]{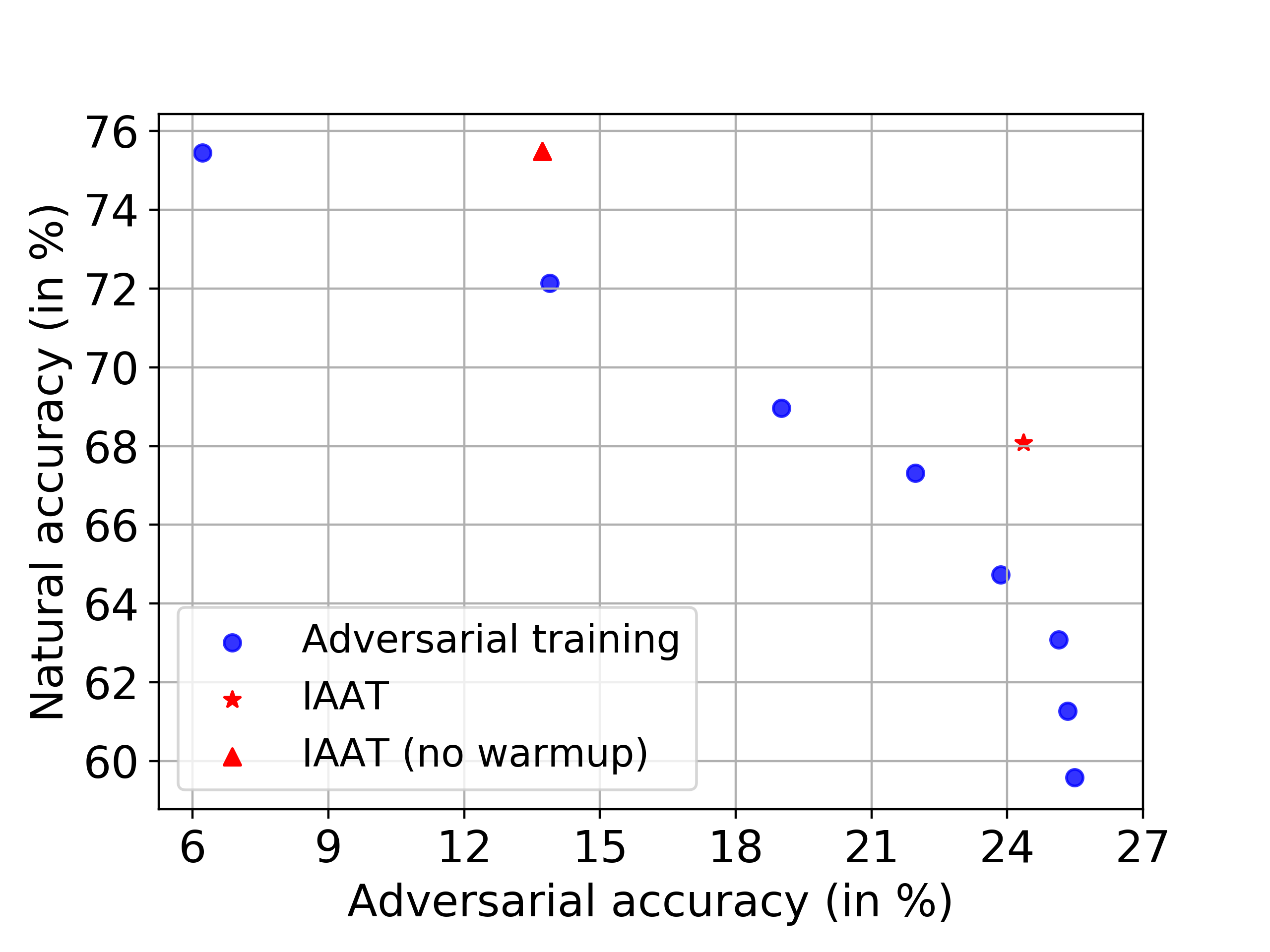}
    \caption{CIFAR-100}
    \label{fig:tradeoff_CIFAR100}
  \end{subfigure}
  \caption{Tradeoffs between accuracy and robustness: Each blue dot denotes an adversarially trained model with a different $\epsilon$. Models trained using instance adaptive adversarial training are shown in red. Adaptive training breaks through the Pareto frontier achieved by plain adversarial training with a fixed $\epsilon$. }
\end{figure}

\paragraph{Beating the robustness-accuracy tradeoff: } In adversarial training, the perturbation radius $\epsilon$ is a hyper-parameter. Training models with varying $\epsilon$ produces a robustness-accuracy tradeoff curve - models with small training $\epsilon$ achieve better natural accuracy and poor adversarial robustness, while models trained on large $\epsilon$ have improved robustness and poor natural accuracy. To generate this tradeoff, we perform adversarial training with $\epsilon$ in the range $\{1/255, 2/255, \hdots 8/255\}$. Instance adaptive adversarial training is then compared with respect to this tradeoff curve in Fig.~\ref{fig:tradeoff_CIFAR10}, \ref{fig:tradeoff_CIFAR100}. Two versions of IAAT are reported - with and without a warmup phase. In both versions, we clearly achieve an improvement over the accuracy-robustness tradeoff. Use of the warmup phase helps retain robustness with a drop in natural accuracy compared to its no-warmup counterpart.
\paragraph{Clean accuracy improves for a fixed level of robustness: } On CIFAR-10, as shown in Table.~\ref{tab:CIFAR10}, we observe that our instance adaptive adversarial training algorithm achieves similar adversarial robustness as the adversarial training baseline. However, the accuracy on clean test samples increases by $4.06 \%$ for Resnet-18 and $4.49 \%$ for WideResnet-32-10. We also observe that the adaptive training algorithm improves robustness to unseen image corruptions. This points to an improvement in overall generalization ability of the network. On CIFAR-100 (Table.~\ref{tab:CIFAR100}), the performance gain in natural test accuracy further increases - $8.79 \%$ for Resnet-18, and $9.22 \%$ for Wideresnet-32-10. The adversarial robustness drop is marginal.


\begin{table}[t]
\caption{Robustness experiments on CIFAR-10. PGD attacks are generated with $\epsilon=8$. PGD$_{10}$ and PGD$_{100}$ attacks are generated with $5$ random restarts, while PGD$_{1000}$ attacks are generated with $2$ random restarts}
\label{tab:CIFAR10}
\begin{center}
\begin{tabular}{|c|c|c|c|c|c|c|}
\hline
\multicolumn{1}{|c|}{Method}  & \multicolumn{1}{c|}{Natural} & \multicolumn{3}{c|}{Whitebox acc. (in $\%$)} & \multicolumn{1}{c|}{Transfer (in $\%$)} & \multicolumn{1}{c|}{Corruption} 
\\  
 & acc. (in $\%$) & PGD$_{10}$ & PGD$_{100}$ & PGD$_{1000}$ & acc. (PGD$_{1000}$) & acc. (in $\%$) \\
 \hline 
 \multicolumn{7}{|c|}{ \textit{Resnet-18}} \\ \hline
Clean               & 94.21 & 0.02 & 0.00 & 0.00 & 3.03 & 72.71 \\
Adversarial         & 83.20 & 43.79 & 42.30 & 42.36 & 59.80 & 73.73 \\ 
IAAT            & 87.26 & 43.08 & 41.16 & 41.16  & 59.87 & 78.82 \\
 \hline 
 \multicolumn{7}{|c|}{ \textit{WideResnet 32-10}} \\ \hline
Clean               & 95.50 & 0.05 & 0.00 & 0.00 &  5.02 & 78.35 \\
Adversarial         & 86.85 & 46.86 & 44.82 & 44.84 & 62.77 & 77.99  \\
IAAT            & 91.34 & 48.53 & 46.50 & 46.54 & 58.20 & 83.13 \\
\hline
\end{tabular}
\end{center}
\end{table}


\begin{table}[t]
\caption{Robustness experiments on CIFAR-100. PGD attacks are generated with $\epsilon=8$. PGD$_{10}$ and PGD$_{100}$ attacks are generated with $5$ random restarts. PGD$_{1000}$ attacks are generated with $2$ random restarts}
\label{tab:CIFAR100}
\begin{center}
\begin{tabular}{|c|c|c|c|c|c|c|c|}
\hline
\multicolumn{1}{|c|}{Method}  & \multicolumn{1}{c|}{Natural} & \multicolumn{3}{c|}{Whitebox acc. (in $\%$)} & \multicolumn{1}{c|}{Transfer acc. (in $\%$)}
\\  
 & acc. (in $\%$) & PGD$_{10}$ & PGD$_{100}$ & PGD$_{1000}$ & PGD$_{1000}$ \\
 \hline 
 \multicolumn{6}{|c|}{ \textit{Resnet-18}} \\ \hline
Clean               & 74.88 & 0.02 & 0.00 & 0.01 & 1.81  \\
Adversarial         & 55.11 & 20.69 & 19.68 & 19.91 & 35.57  \\ 
IAAT            & 63.90 & 18.50 & 17.10 & 17.11 & 35.74 \\
 \hline 
 \multicolumn{6}{|c|}{ \textit{WideResnet 32-10}} \\ \hline
Clean               & 79.91 & 0.01 & 0.00 & 0.00 & 1.20 \\
Adversarial         & 59.58 & 26.24 & 25.47 & 25.49 & 38.10  \\
IAAT            & 68.80 & 26.17 & 24.22 & 24.36 & 35.18 \\
\hline
\end{tabular}
\end{center}
\end{table}

\begin{figure}
  \centering
  \begin{subfigure}[b]{0.45\textwidth}
    \includegraphics[width=\textwidth]{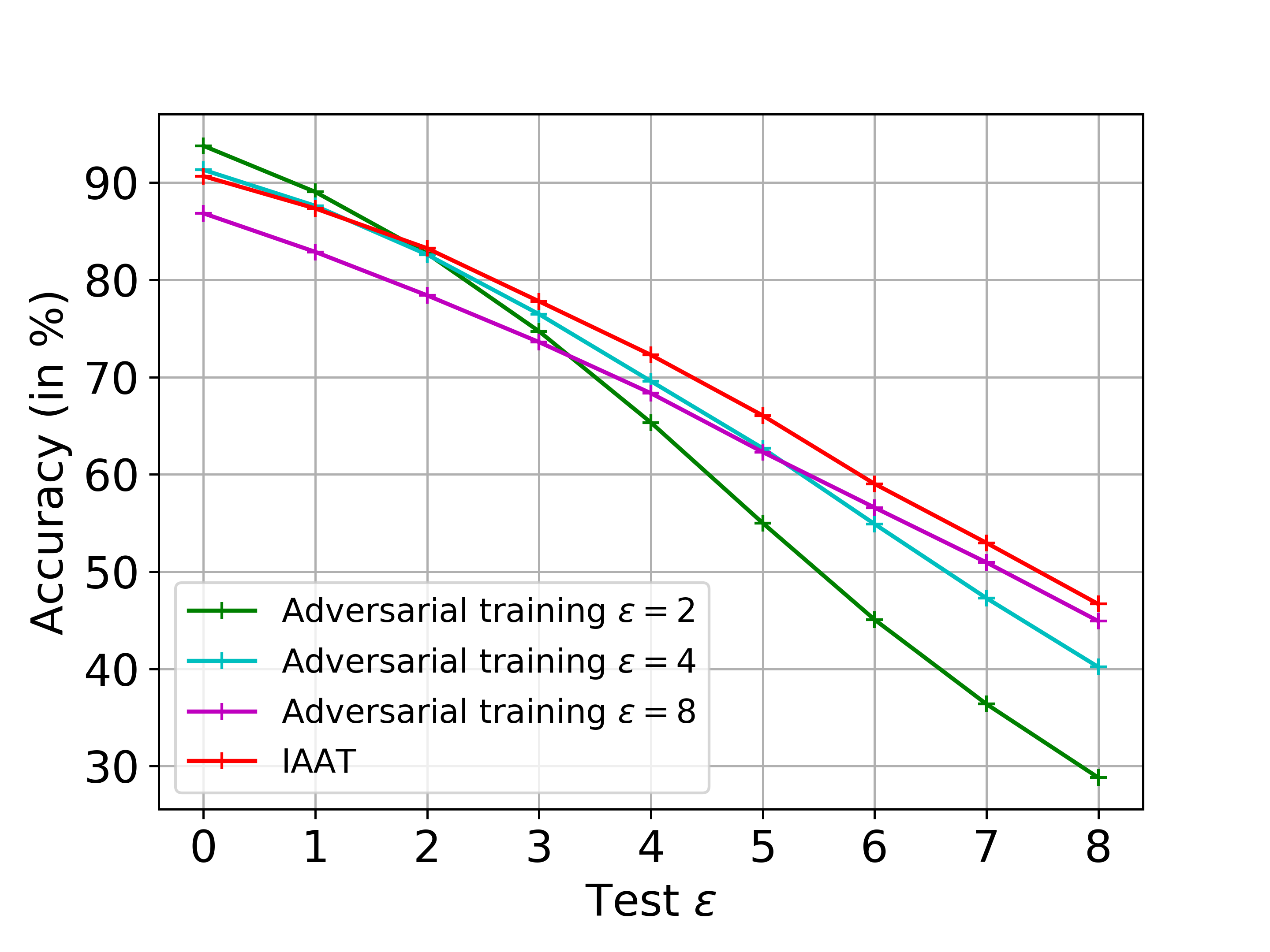}
    \caption{CIFAR-10}
    \label{fig:test_eps_sweep_CIFAR10}
  \end{subfigure}
  \begin{subfigure}[b]{0.45\textwidth}
    \includegraphics[width=\textwidth]{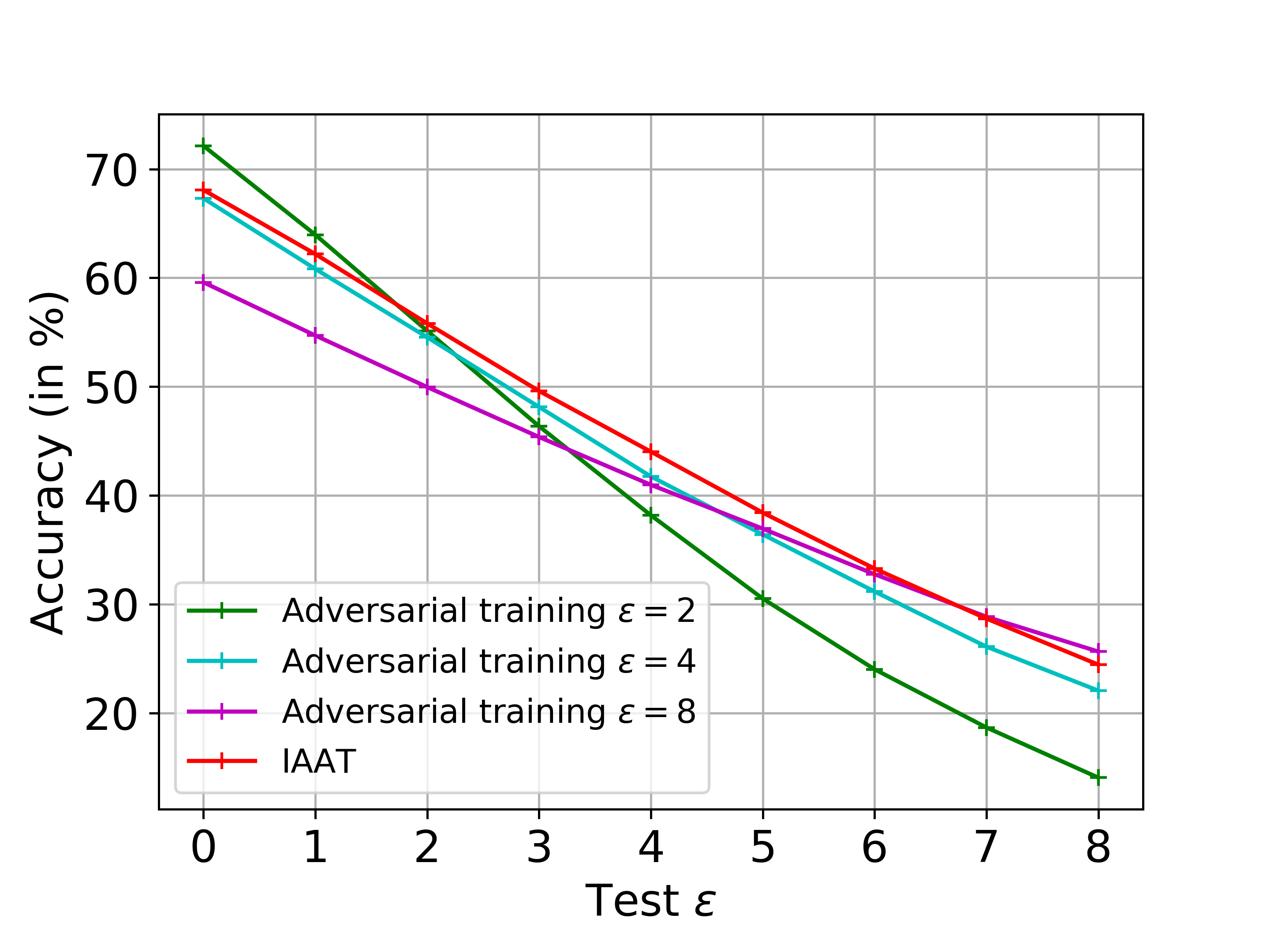}
    \caption{CIFAR-100}
    \label{fig:test_eps_sweep_CIFAR100}
  \end{subfigure}
  \caption{Plot of adversarial robustness over a sweep of test $\epsilon$}
\end{figure}

\paragraph{Maintaining performance over a range of test $\epsilon$: }
Next, we plot adversarial robustness over a sweep of $\epsilon$ values used to craft attacks at test time.  Fig.~\ref{fig:test_eps_sweep_CIFAR10}, \ref{fig:test_eps_sweep_CIFAR100} shows an adversarial training baseline with $\epsilon=8$ performs well at high $\epsilon$ regimes and poorly at low $\epsilon$ regimes. On the other hand, adversarial training with $\epsilon=2$ has a reverse effect, performing well at low $\epsilon$ and poorly at high $\epsilon$ regimes. Our instance adaptive training algorithm maintains good performance over all $\epsilon$ regimes, achieving slightly less performance than the $\epsilon=2$ model for small test $\epsilon,$ and dominating all models for larger test $\epsilon.$

\paragraph{Interpretability of $\epsilon$: }
We find that the values of $\epsilon_i$ chosen by our adaptive algorithm correlate well with our own human concept of class ambiguity. Figure \ref{fig:sample_vis_main} (and Figure \ref{fig:sample_vis_appendix} in Appendix~\ref{app:sample_vis}) shows that a sampling of images that receive small $\epsilon_i$ contains many ambiguous images, and these images are perturbed into a (visually) different class using $\epsilon=16.$  In contrast, images that receive a large $\epsilon_i$ have a visually definite class, and are not substantially altered by an $\epsilon=16$ perturbation. 

\subsection{Imagenet}

Following the protocol introduced in \cite{Xie_2019_feature}, we attack Imagenet models using random targeted attacks instead of untargeted attacks as done in previous experiments. During training, adversarial attacks are generated using $30$ steps of PGD. As a baseline, we use adversarial training with a fixed $\epsilon$ of $16/255$. This is the setting used in \cite{Xie_2019_feature}. Adversarial training on Imagenet is computationally intensive. To make training practical, we use distributed training with synchronized SGD on $64 / 128$ GPUs. More implementation details can be found in Appendix~\ref{abl:implementation}. 

At test time, we evaluate the models on clean test samples and on whitebox adversarial attacks with $\epsilon = \{ 4, 8, 12, 16\}$. PGD-$1000$ attacks are used. Additionally, we also report normalized mean corruption error (mCE), an evaluation metric introduced in \cite{hendrycks2018benchmarking} to test the robustness of neural networks to image corruptions. This metric reports mean classification error of different image corruptions averaged over varying levels of degradation. Note that while accuracies are reported for natural and adversarial robustness, mCE reports classification errors, so lower numbers are better. 

Our experimental results are reported in Table.~\ref{tab:Imagenet}. We observe a huge drop in natural accuracy for adversarial training ($25 \%$, $22 \%$ and $20 \%$ drop for Resnet-50, 101 and 152 respectively). Adaptive adversarial training significantly improves the natural accuracy - we obtain a consistent performance gain of $10+ \%$ on all three models over the adversarial training baseline. On whitebox attacks, IAAT outperforms the adversarial training baseline on low $\epsilon$ regimes, however a drop of $~ 13\%$ is observed at high $\epsilon$'s ($\epsilon = 16$). On the corruption dataset, our model consistently outperforms adversarial training.


\begin{table}[t]
\caption{Robustness experiments on Imagenet. All adversarial attacks are generated with PGD-1000. ($\uparrow$) indicates higher numbers are better, while ($\downarrow$) indicates lower numbers are better}
\label{tab:Imagenet}
\begin{center}
\begin{tabular}{|c|c|c|c|c|c|c|}
\hline
\multicolumn{1}{|c|}{Method}  & \multicolumn{1}{c|}{Natural} & \multicolumn{4}{c|}{Whitebox acc. (in $\%$) ($\uparrow$) } & \multicolumn{1}{c|}{Corruption} 
\\  
 & acc. (in $\%$) ($\uparrow$) & $\eps = 4$ & $\eps=8$ & $\eps=12$ & $\eps=16$ & mCE ($\downarrow$)  \\
 \hline 
 \multicolumn{7}{|c|}{ \textit{Resnet-50}} \\ \hline
Clean training         & 75.80 & 0.64 & 0.18 & 0.00 & 0.00 & 76.69 \\
Adversarial training   & 50.99 & 50.89 & 49.11 & 44.71 & 35.82 & 95.48  \\
IAAT   & 62.71 & 61.52 & 54.63 & 39.90 & 22.72 & 85.21 \\
 \hline 
 \multicolumn{7}{|c|}{ \textit{Resnet-101}} \\ \hline
Clean training         & 77.10 & 0.83 & 0.12 & 0.00 & 0.00 & 70.37 \\
Adversarial training   & 55.42 & 55.11 & 53.07 & 48.35 & 39.08 & 91.45 \\
IAAT   & 65.29 & 63.83 & 56.62 & 41.51 & 23.91 & 79.52 \\
\hline
 \multicolumn{7}{|c|}{ \textit{Resnet-152}} \\ \hline
Clean training         & 77.60 & 0.57 & 0.08 & 0.00 & 0.00 & 69.27 \\
Adversarial training   & 57.26 & 56.77 & 54.75 & 49.86 & 40.40 & 89.31 \\
IAAT   & 67.44 & 65.97 & 59.28 & 45.01 & 27.85 & 78.53 \\
\hline
\end{tabular}
\end{center}
\end{table}


\section{Ablation experiments}\label{sec:ablation}

\subsection{Effect of warmup}

\begin{table}[t]
\caption{Ablation: Effect of warmup on CIFAR-10}
\label{tab:abl_warm_CIFAR10}
\begin{center}
\begin{tabular}{|c|c|c|c|c|c|c|}
\hline
\multicolumn{1}{|c|}{Method}  & \multicolumn{1}{c|}{Natural} & \multicolumn{3}{c|}{Whitebox acc. (in $\%$)} & \multicolumn{1}{c|}{Transfer acc.($\%$)} & \multicolumn{1}{c|}{Corruption} 
\\  
 & acc. ($\%$) & PGD$_{10}$ & PGD$_{100}$ & PGD$_{1000}$  & PGD$_{1000}$ & acc. (in $\%$) \\
 \hline 
 \multicolumn{7}{|c|}{ \textit{Resnet-18}} \\ \hline
IAAT (no warm)  & 89.62 & 40.55 & 38.15 & 38.08 &  58.89 & 81.10 \\
IAAT (warm)     & 87.26 & 43.08 & 41.16 & 41.16  &  59.87 & 78.82 \\
 \hline 
 \multicolumn{7}{|c|}{ \textit{WideResnet 32-10}} \\ \hline
IAAT (no warm)  & 92.62 & 45.12 & 41.08 & 41.11 &  53.08 & 84.92  \\
IAAT (warm)     & 90.67 & 48.53 & 46.50 & 46.54 &  58.20 & 83.13 \\
\hline
\end{tabular}
\end{center}
\end{table}

\begin{table}[t]
\caption{Ablation: Effect of warmup on CIFAR-100}
\label{tab:abl_warm_CIFAR100}
\begin{center}
\begin{tabular}{|c|c|c|c|c|c|}
\hline
\multicolumn{1}{|c|}{Method}  & \multicolumn{1}{c|}{Natural} & \multicolumn{3}{c|}{Whitebox acc. (in $\%$)} & \multicolumn{1}{c|}{Transfer acc.($\%$)}
\\  
 & acc. (in $\%$) & PGD$_{10}$ & PGD$_{100}$ & PGD$_{1000}$ & PGD$_{1000}$ \\
 \hline 
 \multicolumn{6}{|c|}{ \textit{Resnet-18}} \\ \hline
Adaptive (no warm)  & 68.34 & 14.76 & 13.29 & 13.30 &  32.39 \\
Adaptive (warm)     & 63.90 & 18.50 & 17.10 & 17.11 &  35.74 \\
 \hline 
 \multicolumn{6}{|c|}{ \textit{WideResnet 32-10}} \\ \hline
Adaptive (no warm)  & 75.48 & 18.14 & 13.78 & 13.71 &  24.00  \\
Adaptive (warm)     & 68.80 & 26.17 & 24.22 & 24.36 &  35.18 \\
\hline
\end{tabular}
\end{center}
\end{table}

In this section, we study the effect of using a warmup phase in adaptive adversarial training. Recall from Section~\ref{sec:method} that during warmup, adversarial training is performed with uniform norm-bound constraints. Once the warmup phase ends, we switch to instance adaptive training. From Table~\ref{tab:abl_warm_CIFAR10} and \ref{tab:abl_warm_CIFAR100}, we observe that when warmup is used, adversarial robustness improves with a small drop in natural accuracy. The improvement in robustness is more pronounced in the CIFAR-100 dataset. However, as shown in Fig.~\ref{fig:tradeoff_CIFAR10} and \ref{fig:tradeoff_CIFAR100}, both these settings improve the accuracy-robustness tradeoff.

\begin{figure}
  \centering
  \begin{subfigure}[b]{0.45\textwidth}
    \includegraphics[width=\textwidth]{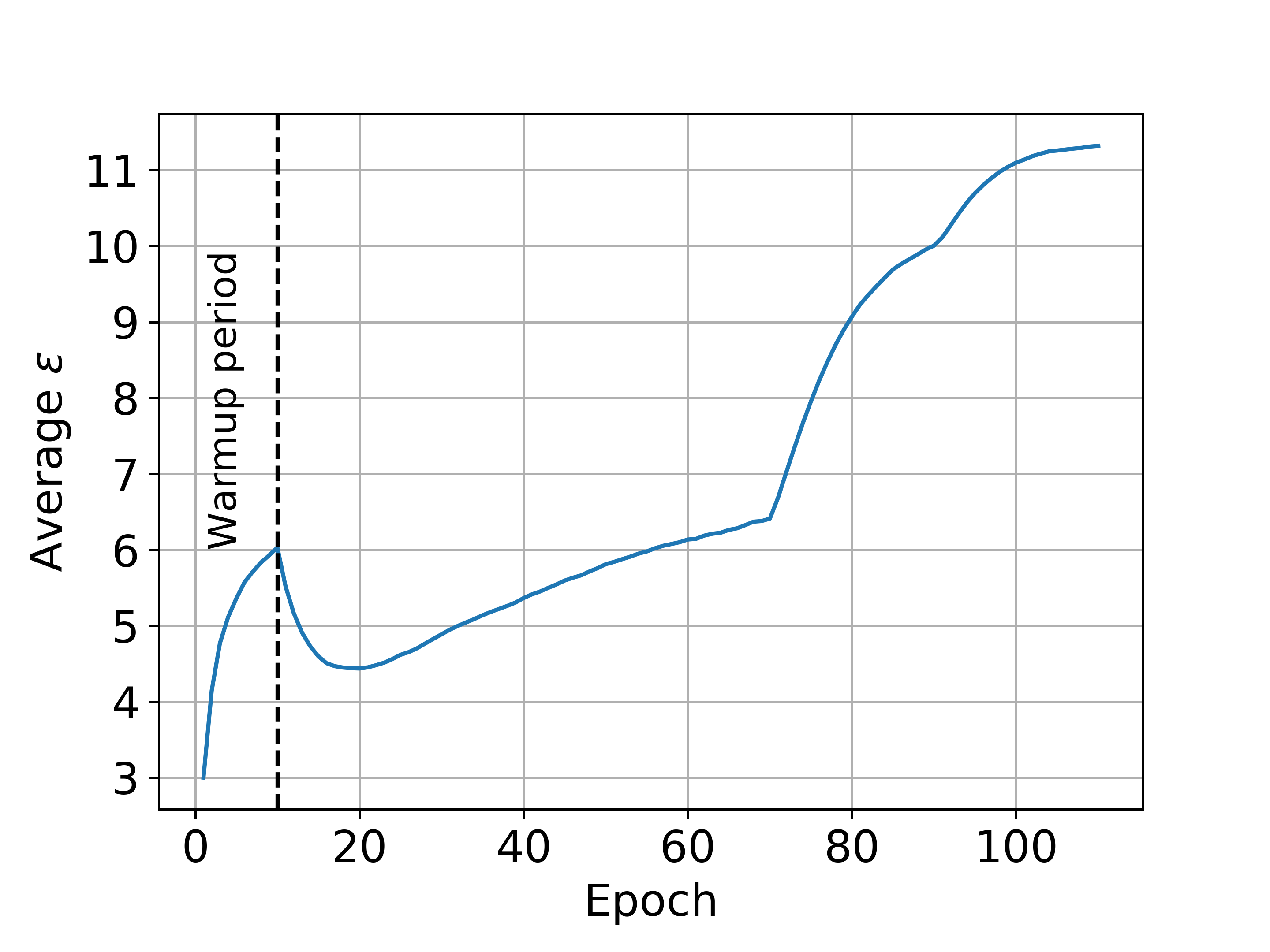}
    \caption{Average $epsilons$}
    \label{fig:eps_avg}
  \end{subfigure}
  \begin{subfigure}[b]{0.45\textwidth}
    \includegraphics[width=\textwidth]{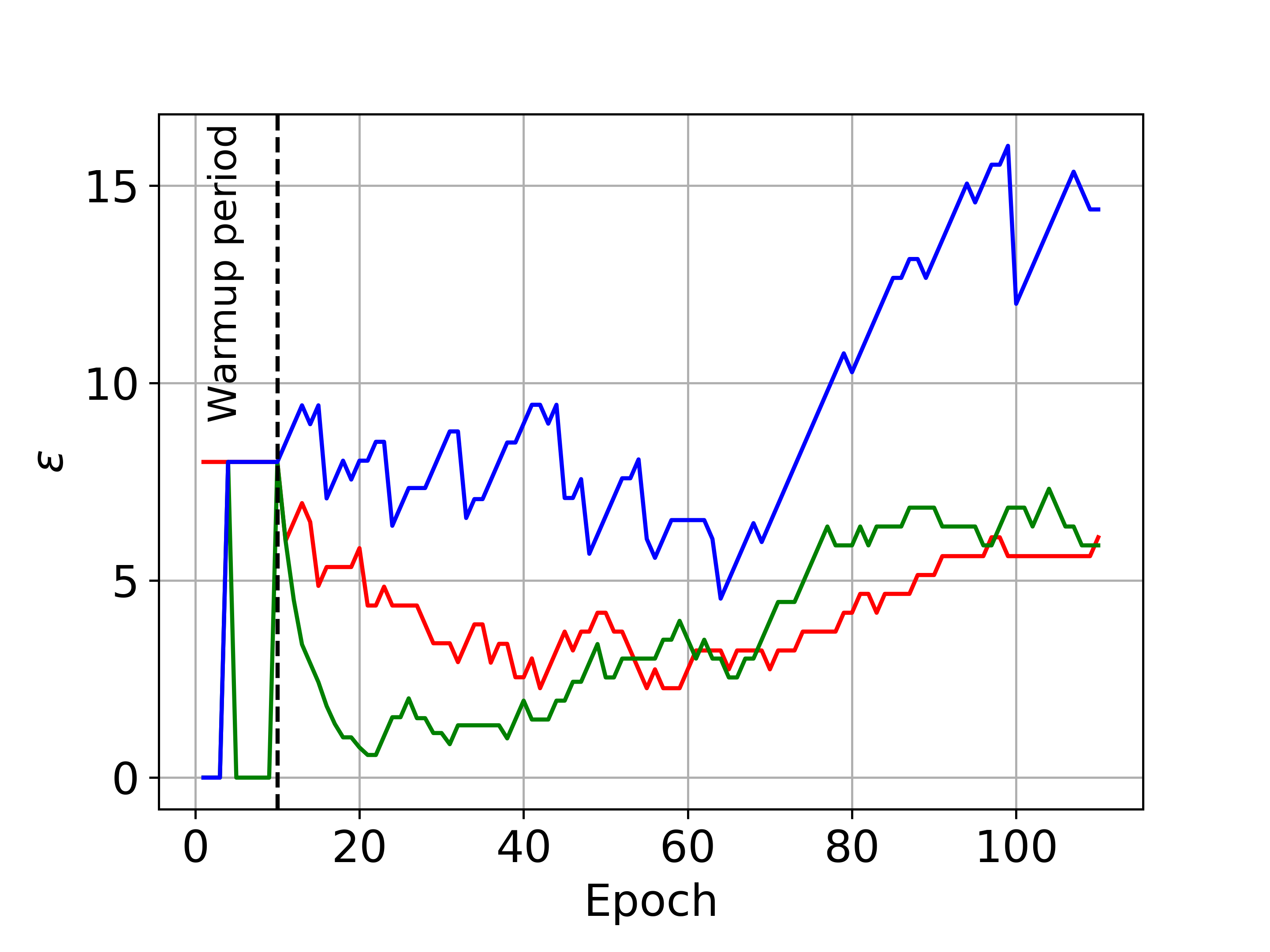}
    \caption{Individual $\epsilon$ for $3$ random samples}
    \label{fig:eps_ind}
  \end{subfigure}
  \caption{Visualizing $\epsilon$ progress of instance adaptive adversarial trianing. Plot on the left shows average $\epsilon$ of samples over epochs, while the plot on the right shows $\epsilon$ progress of three randomly chosen samples.}
\end{figure}

\subsection{Visualizing $\epsilon$ progress}
Next, we visualize the evolution of $\epsilon$ over epochs in adaptive adversarial training. A plot showing the average $\epsilon$ growth, along with the $\epsilon$ progress of $3$ randomly picked samples are shown in Fig.~\ref{fig:eps_avg} and \ref{fig:eps_ind}. We observe that average $\epsilon$ converges to around $11$, which is higher than the default setting of $\epsilon=8$ used in adversarial training. Also, each sample has a different $\epsilon$ profile - for some, $\epsilon$ increases well beyond the commonly use radius of $\epsilon=8$, while for others, it converges below it. In addition, a plot showing the histogram of $\epsilon$'s at different snapshots of training is shown in Fig.~\ref{fig:eps_hist}. We observe an increase in spread of the histogram as the training progresses.


\section{Conclusion}\label{sec:conclusion}

In this work, we focus on improving the robustness-accuracy tradeoff in adversarial training. We first show that realizable robustness is a sample-specific attribute: samples close to the decision boundary can only achieve robustness within a small $\epsilon$ ball, as they contain samples from a different class beyond this radius. On the other hand samples far from the decision boundary can be robust on a relatively large perturbation radius. Motivated by this observation, we develop \textit{instance adaptive adversarial training}, in which label consistency constraints are imposed within sample-specific perturbation radii, which are in-turn estimated. Our proposed algorithm has empirically been shown to improve the robustness-accuracy tradeoff in CIFAR-10, CIFAR-100 and Imagenet datasets.  

\section{Acknowledgements}
Goldstein and Balaji were supported in part by the DARPA GARD program, DARPA QED for RML, DARPA Lifelong Learning Machines,  the DARPA Young Faculty Award program, the AFOSR MURI program, and the National Science Foundation.

\bibliography{iclr2020_conference}
\bibliographystyle{iclr2020_conference}


\newpage
\appendix
\section{Appendix}

\subsection{Comparison with Mixup}

A recent paper that addresses the problem of improving natural accuracy in adversarial training is mixup adversarial training~\citep{lamb2019interpolated}, where adversarially trained models are optimized using mixup loss instead of the standard cross-entropy loss. In this paper, natural accuracy was shown to improve with no drop in adversarial robustness. However, the robustness experiments were not evaluated on strong attacks (experiments were reported only on PGD-20). We compare our implementation of mixup adversarial training with IAAT on stronger attacks in Table.~\ref{tab:comparison_mixup}. We observe that while natural accuracy improves for mixup, drop in adversarial accuracy is much higher than IAAT.

\begin{table}[h]
\caption{Comparison with Mixup}
\label{tab:comparison_mixup}
\begin{center}
\begin{tabular}{|c|c|c|c|c|c|}
\hline
\multicolumn{1}{|c|}{Method}  & \multicolumn{1}{c|}{Natural} & \multicolumn{3}{c|}{Whitebox acc. (in $\%$)} & \multicolumn{1}{c|}{Transfer attack (in $\%$)} 
\\  
 & acc. (in $\%$) & PGD$_{10}$ & PGD$_{100}$ & PGD$_{1000}$ & PGD$_{1000}$  \\
 \hline 
 \multicolumn{6}{|c|}{ \textit{Resnet-18}} \\ \hline
Mixup               & 89.47 & 42.60 & 38.42 & 38.49 & 59.48  \\
IAAT            & 87.26 & 43.08 & 41.16 & 41.16  & 59.87 \\
 \hline 
 \multicolumn{6}{|c|}{ \textit{WideResnet 32-10}} \\ \hline
Mixup               & 92.57 & 45.01 & 36.6 & 36.44 & 63.57  \\             
IAAT            & 90.67 & 48.53 & 46.50 & 46.54 & 58.20 \\
\hline
\end{tabular}
\end{center}
\end{table}

\section{Sample visualization}\label{app:sample_vis}
\begin{figure*}
\centering
\includegraphics[width=\textwidth]{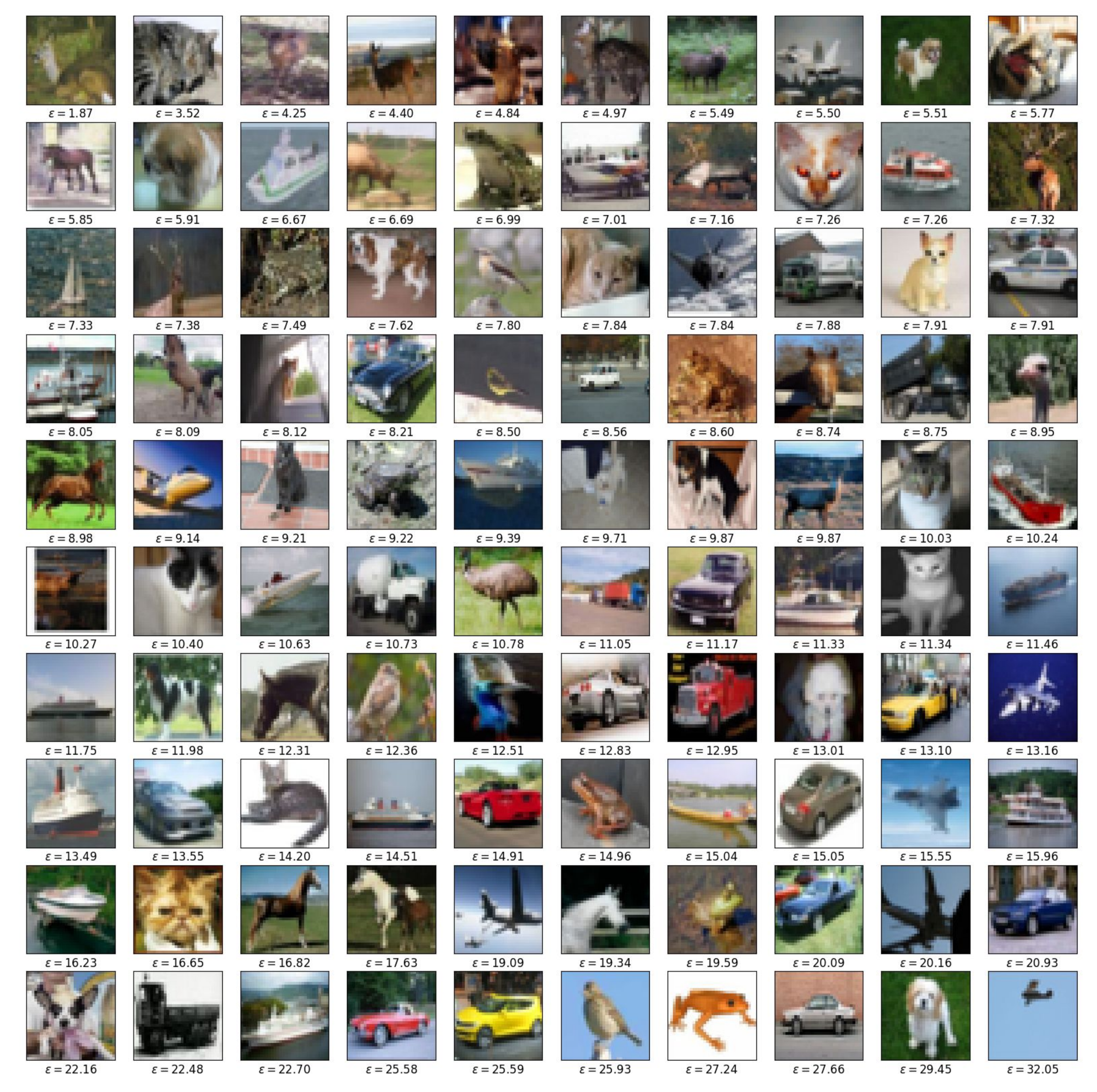}
\caption{Visualizing training samples with their corresponding perturbation. All $\epsilon$ live in the range $[0, 255]$}
\label{fig:sample_vis_full} 
\end{figure*}

A visualization of samples from CIFAR-10 dataset with the corresponding $\epsilon$ value assigned by IAAT is shown in Figure.~\ref{fig:sample_vis_full}. We observe that samples for which low $\epsilon$'s are assigned are visually confusing (eg., top row of Figure.~\ref{fig:sample_vis_full}), while samples with high $\epsilon$ distinctively belong to one class.

\begin{figure*}
\centering
\includegraphics[width=\textwidth]{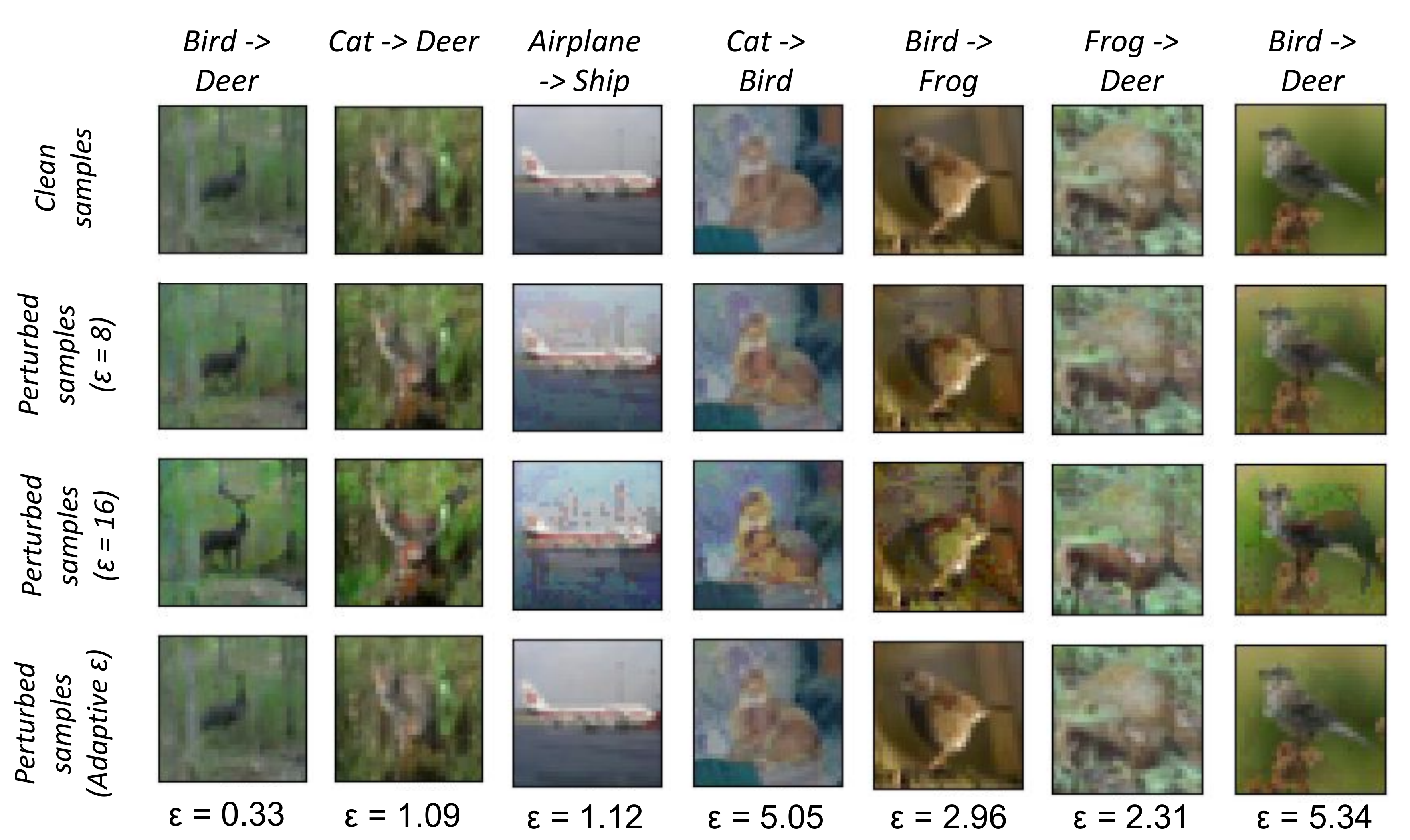}
\caption{Visualizations of samples for which low $\epsilon$'s are assigned by instance adaptive adversarial training. These samples are close to the decision boundary and change class when perturbed with $\epsilon \geq 8$. Perturbing them with $\epsilon$ assigned by IAAT retains the class information.}
\label{fig:sample_vis_appendix} 
\end{figure*}

In addition, we also show more visualizations of samples near decision boundary which contain samples from a different class within a fixed $\ell_{\infty}$ ball in Figure.~\ref{fig:sample_vis_appendix}. The infeasibility of label consistency constraints within the commonly used perturbation radius of $\ell_{\infty} = 8$ is apparent in this visualization. Our algorithm effectively chooses an appropriate $\epsilon$ that retains label information within the chosen radius.

\begin{figure*}
\centering
\includegraphics[width=\textwidth]{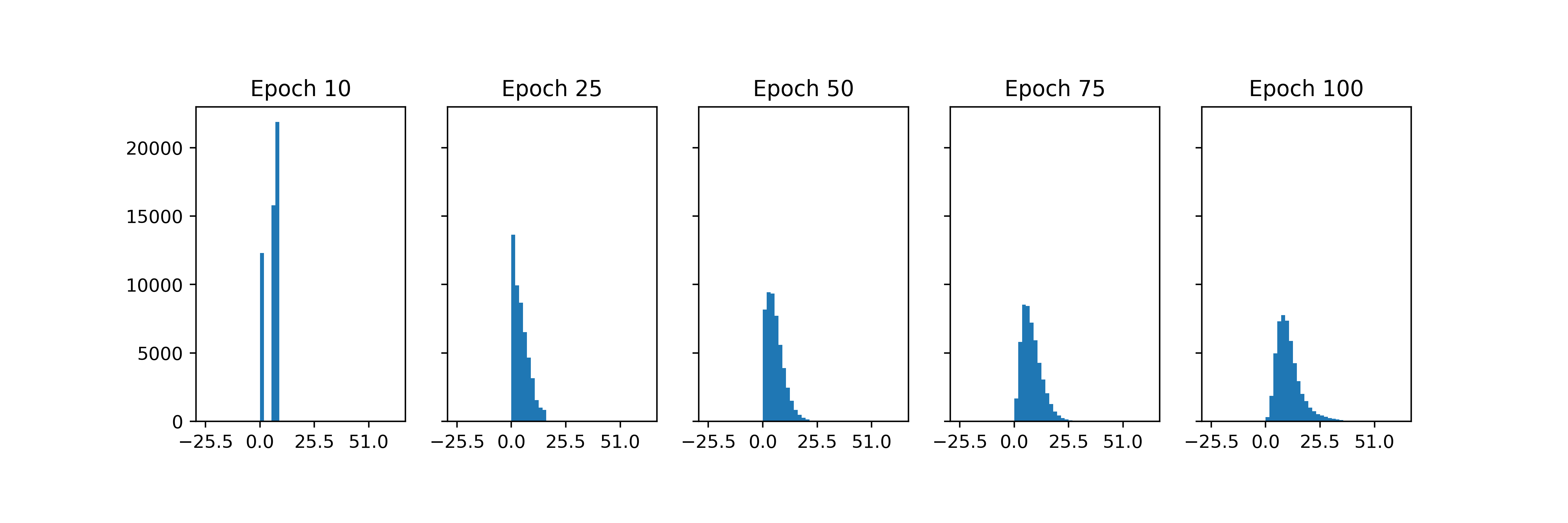}
\caption{Histogram of $\epsilon$ of training samples at different training epochs}
\label{fig:eps_hist} 
\end{figure*}

\section{Imagenet sweep over PGD iterations}

\begin{figure}
\centering
\includegraphics[width=0.7\textwidth]{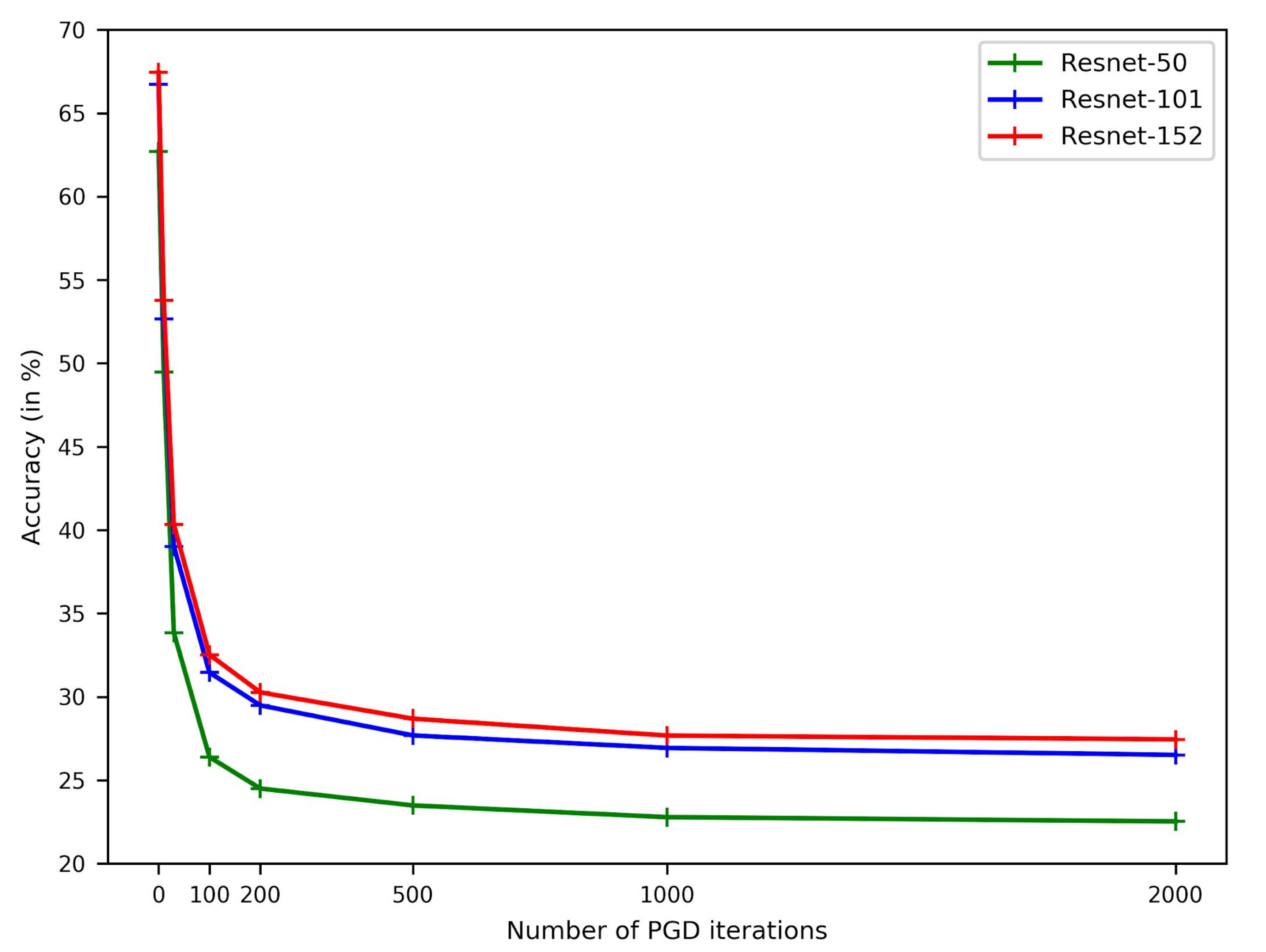}
\caption{Imagenet robustness of IAAT over the number of PGD iterations}
\label{fig:imagenet_saturation} 
\end{figure}

Testing against a strong adversary is crucial to assess the true robustness of a model. A popular practice in adversarial robustness community is to attack models using PGD with many attack iterations~\citep{Xie_2019_feature}. So, we test our instance adaptive adversarially trained models on a sweep of PGD iterations for a fixed $\epsilon$ level. Following ~\citep{Xie_2019_feature}, we perform the sweep upto $2000$ attack steps fixing $\epsilon=16$. The resulting plot is shown in Figure.~\ref{fig:imagenet_saturation}. For all three Resnet models, we observe a saturation in adversarial robustness beyond 500 attack iterations.

\section{Implementation details}\label{abl:implementation}

\subsection{CIFAR}

On CIFAR-10 and CIFAR-100 datasets, our implementation follows the standard \textit{adversarial training} setting used in \cite{madry2018towards}. During training, adversarial examples are generated using PGD-10 attacks, which are then used to update the model. All hyperparameters we used are tabulated in Table.~\ref{tab:hyp_CIFAR}.

\begin{table*}
\caption{Hyper-parameters for experiments on CIFAR-10 and CIFAR-100}
\label{tab:hyp_CIFAR}
\centering
\begin{tabular}{c|c|c}
\hline
Hyperparameters & Resnet-18 & WideResnet-32-10 \\
\hline
\hline
Optimizer & SGD & SGD\\
Start learning rate & 0.1 & 0.1 \\
Weight decay & 0.0002 & 0.0005 \\
Number of epochs trained & 200 & 110 \\
Learning rate annealing & Step decay & Step decay \\
Learning rate decay steps & [80, 140, 170] & [70, 90, 100] \\
Learning rate decay factor & 0.1 & 0.2 \\
Batch size & 128 & 128 \\
\hline
Warmup period & 5 epochs & 10 epochs \\
$\epsilon$ used in warmup ($\epsilon_w$) & 8 & 8 \\
Discretization $\gamma$ & 1.9 & 1.9 \\
Exponential averaging factor $\beta$ & 0.1 & 0.1 \\
\hline
\hline
\multicolumn{3}{c}{Attack parameters during training} \\
\hline
Attack steps & 10 & 10 \\
Attack $\epsilon$ (for adv. training only) & 8 & 8 \\
Attack learning rate & 2/255 & 2/255 \\
\hline
\end{tabular}
\end{table*}

\subsection{Imagenet}

For Imagenet implementation, we mimic the setting used in \cite{Xie_2019_feature}. During training, adversaries are generated with PGD-30 attacks. This is computationally expensive as every training update is followed by $30$ backprop iterations to generate the adversarial attack. To make training feasible, we perform distributed training using synchronized SGD updates on 64 / 128 GPUs. We follow the training recipe introduced in \cite{goyal2017imagenet1hr} for large batch training. Also, during training, adversarial attacks are generated with FP-16 precision. However, in test phase, we use FP-32. 

We further use two more tricks to speed-up instance adaptive adversarial training: (1) A weaker attacker(PGD-10) is used in the algorithm for selecting $\epsilon$ (Alg.~\ref{alg:epsilon_select}). (2) After $\epsilon_i$ is selected per Alg.~\ref{alg:epsilon_select}, we clip it with a lower-bound i.e., $\epsilon_i \gets max(\epsilon_i, \epsilon_{lb})$. $\epsilon_{lb}=4$ was used in our experiments. 

Hyperparameters used in our experiments are reported in Table~\ref{tab:hyp_Imagenet}. All our models were trained on PyTorch.

\begin{table*}
\caption{Hyper-parameters for experiments on Imagenet}
\label{tab:hyp_Imagenet}
\centering
\begin{tabular}{c|c}
\hline
Hyperparameters & Imagenet \\
\hline
\hline
Optimizer & SGD \\
Start learning rate & 0.1 $\times$ (effective batch size / 256) \\
Weight decay & 0.0001 \\
Number of epochs trained & 110 \\
Learning rate annealing & Step decay with LR warmup \\
Learning rate decay steps & [35, 70, 95] \\
Learning rate decay factor & 0.1 \\
Batch size & 32 per GPU \\
\hline
Warmup period & 30 epochs \\
$\epsilon$ used in warmup ($\epsilon_w$) & 16 \\
Discretization $\gamma$ & 4 \\
Exponential averaging factor $\beta$ & 0.1 \\
\hline
\hline
\multicolumn{2}{c}{Attack parameters during training} \\
\hline
Attack steps & 30 \\
Attack $\epsilon$ (for adv. training only) & 16 \\
Attack learning rate & 1/255 \\
\hline
\end{tabular}
\end{table*}

Resnet-50 model was trained on 64 Nvidia V100 GPUs, while Resnet-101 and Resnet-152 models were trained on 128 GPUs. Time taken for instance adaptive adversarial training for all models is reported in Table.~\ref{tab:time_imagenet}. 

\begin{table*}
\caption{Training time for Imagenet experiments}
\label{tab:time_imagenet}
\centering
\begin{tabular}{c|c|c}
\hline
Model & Number of GPUs used & Training time \\
\hline
\hline
Resnet-50 & 64 & 92 hrs \\
Resnet-101 & 128 & 78 hrs \\
Resnet-152 & 128 & 94 hrs \\
\hline
\end{tabular}
\end{table*}

\end{document}